\documentclass[conference]{IEEEtran}
\IEEEoverridecommandlockouts

\usepackage{cite}
\usepackage{amsmath,amssymb,amsfonts}
\usepackage{algorithmic}
\usepackage{graphicx}
\usepackage{textcomp}
\usepackage{xcolor}
\def\BibTeX{{\rm B\kern-.05em{\sc i\kern-.025em b}\kern-.08em
    T\kern-.1667em\lower.7ex\hbox{E}\kern-.125emX}}

\begin{document}

\title{News Reporter: A Multi-lingual LLM Framework for Broadcast T.V News}

\author{\IEEEauthorblockN{Tarun Jain*\thanks{*Equal contribution.}}
\IEEEauthorblockA{
\textit{AI Planet}\\
tarun@aiplanet.com}
\and
\IEEEauthorblockN{Yufei Gao*}
\IEEEauthorblockA{
\textit{East China Normal University}\\
10215501422@stu.ecnu.edu.cn}
\and
\IEEEauthorblockN{Sridhar Vanga, Karan Singla}
\IEEEauthorblockA{
\textit{Whissle AI}\\
\{svanga, ksingla\}@whissle.ai}
}

\maketitle

\begin{abstract}

Large Language Models (LLMs) have fast become an essential tools to many conversational chatbots due to their ability to provide coherent answers for varied queries. Datasets used to train these LLMs are often a mix of generic and synthetic samples, thus lacking the verification needed to provide correct and verifiable answers for T.V. News.

We collect and share a large collection of QA pairs extracted from transcripts of news recordings from various news-channels across the United States. Resultant QA pairs are then used to fine-tune an off-the-shelf LLM model. Our model surpasses base models of similar size on several open LLM benchmarks. We further integrate and propose a RAG method to improve contextualization of our answers and also point it to a verifiable news recording. 

\end{abstract}
\section{Introduction}




Large Language Models (LLMs) have fast become essentials components of many conversational chatbots capable to answer queries on a large range of topics. Many industrial LLMs are deployed as general purpose search engines to provide answers for everyday need. However, these LLMs are trained on mix of real and synthetic data \cite{dubey2024llama}, which makes them unreliable when it comes to answering questions about news, be it from newspaper articles or T.V News. 

T.V news, which is also the topic of this paper, becomes a central topic on which users are interested in asking queries \cite{diaz2009integration}. While most industrial LLM based assistant (ChatGPT, Claude, Gemini), answer news queries but without any verifiable sources. Few LLM based solutions like Perplexity, focuses on using RAG (Retrieval based generation) to also find and point news articles for user queries. However, reliability of these results remains largely unknown \cite{wang2023surveyfactualitylargelanguage}, as there are no published evaluation results. Moreover, these solutions are limited to newspaper web articles \cite{xiao2023enhancing} and have no access to  broadcast T.V news.

We use a large collection of transcribed broadcast news recordings, extract Question-Answer (QA) pairs and fine-tune an off-the-shelf LLM. Our transcribed recordings are provided by open-source organization Red Hen Lab \footnote{https://www.redhenlab.org/} which has access to video recordings from UCLA news archives \footnote{https://guides.library.ucla.edu/news/broadcast}.  We automatically extract QA pairs using Self-Instruct functional of cloud LLM service providers for an year of transcriptions. In total, we collect and share the resultant 40k QA pairs for English and 20k QA pairs collectively for French, Spanish, German, and Portuguese. These QA pairs are then used to fine-tuning purposes.

We pick an off-the-shelf pretrained LLM model \footnote{https://huggingface.co/microsoft/Phi-3-mini-4k-instruct} as our base model and then fine-tune it using QA pairs collected from broadcast T.V News. Our model \textbf{News-reporter-3B}, which represents the first LLM adapted for real broadcast T.V News exhibits behavioral characteristics akin to a news reporter. We use 11-month data from 2016 for 35+ news channels in the united states and 1-month for evaluation purposes. We evaluate our model on various different LLM benchmarks, and also perform evaluation using a RAG (Retrievel-augmented-generation) pipeline to point to a verifiable for each generated answer.

Fine-tuned model surpasses base models such as Gemma-7B, Llama-2-7B, and Mistral-7B open-source models on Open LLM benchmarking. Results improve further when we use a vectorDB (learnt using 11-month news context) data for better contextualization. While sharing news transcriptions is restricted due to copyright permissions, it is permissible to share vectorized database, allowing our generated answers to be mapped to an actual news source. We perform an in-depth exploration of metrics which are more appropriate for evaluating responses pertaining to news-related content.

Here are the contributions of this paper:
\begin{itemize}
    \item QA pair dataset collected from real broadcast news. We also provide a vector database for an year of news broadcasts to provide verifiable sources for each answer.
    \item A state-of-the-art open-source fine-tuned on real broadcast T.V News transcripts \footnote{https://huggingface.co/RedHenLabs/news-reporter-3b}.
    \item Evaluation results of the effectiveness, correctness of different LLMs on answering news queries.
    
\end{itemize}

\section{Related Work}
In this section, we first discuss reliability of LLMs for answering queries related to news. We then discuss LLMs which are adapted using news articles. We conclude by discussing opportunities and need to fine-tune LLMs for broadcast news.

\subsection{LLMs are unreliable for news}




Large Language Models (LLMs) understand and process text across various languages, including news transcripts. A prominent example is BLOOM, a 176B-parameter open-access multilingual language model created by the BigScience collaboration \cite{le2023bloom}. BLOOM is trained in 46 human languages as well as 13 programming languages. It leverages large-scale textual data from diverse sources such as literature, scientific articles, and news broadcasts. However, despite their impressive capabilities, they face challenges in reliably processing news content. A recent survey study \cite{tonmoy2024comprehensive} suggests that such models remain generic and often rely on synthetic data, limiting their ability to provide reliable answers for news.

\subsection{LLMs are fine-tuned on text news articles}

\cite{xiao2023enhancing} uses contextual fine-tuning to improve LLM performance on news-related tasks. Specifically, the authors propose using LLM to extract multiple structured event patterns from news paragraphs through unsupervised learning. This process involves selecting an overall event type for the news to generalize, then identifying different events from the text and generating event patterns based on each event. Nevertheless, it's important to note that while fine-tuning can improve LLM performance on news tasks, it does not entirely resolve the underlying issues of reliability and accuracy, especially when dealing with current events or broadcast news data.

\subsection{LLMs being fine-tuned for broadcast news}

Large language models (LLMs) are increasingly being fine-tuned for domain-specific tasks like broadcast news transcription and analysis. Building on earlier work such as Adda et al.'s optimization of language models with transcribed television and radio broadcasts \cite{adda1999language}, models like RadioTalk have emerged, offering a vast corpus of 2.8 billion words from 284,000 hours of radio speech \cite{beeferman2019radiotalk}. This enables LLMs to improve transcription accuracy and content analysis for news and talk radio.

Further advancements in instruction fine-tuning allow LLMs to follow structured prompts and generate reliable, citation-based responses \cite{arxiv_instruction_finetuning_2409_00096, arxiv_instruction_survey_2308_10792}. Techniques such as Retrieval-Augmented Generation (RAG) enhance the accuracy of these models by retrieving current and verified information, crucial for broadcast applications where reliability is key \cite{acorn_finetuning_methods, ultimate_guide_finetuning_2408_13296, towards_ai_instruction_finetuning}.

\section{Dataset}

The raw data collection from the UCLA News Archives spanned over a year, with monthly data from news channels, including CNN, MSNBC, FOX News, BBC, and Al Jazeera. This diverse dataset captures news in various languages and regions. Data was processed using Red Hen Lab's audio-processing pipelines, formatted with metadata such as unique IDs, duration, resolution, language, and timestamps for precise identification. The collected metadata provides essential context to improve the relevance and accuracy of information retrieved by the RAG pipeline.

\subsection{Data Collection}

\begin{table}
\centering
\begin{tabular}{|l|l|l|l|}
\hline
Language & \#Channels & \#Recordings & \#Hours \\ \hline
English  &       21          &      68325        &   61752      \\ \hline
Spanish  &        7         &      1242       &   894      \\ \hline
French  &         2        &       514      &    327     \\ \hline
German  &         3       &       1849     &     555    \\ \hline
Portuguese  &     2            &     777         &   760      \\ \hline
\end{tabular}
\caption{Raw data statistics}
\label{tab:raw-data-statistics}
\end{table}

Table \ref{tab:raw-data-statistics} shows statics of raw data collected for one year 2016. The dataset is multilingual, reflecting the global nature of news media. We only keep recordings for English, French, German, Spanish, and Portuguese. The dataset collected is noisy and needs preprocessing to make it suitable for fine-tuning a pre-trained LLM checkpoint and for building a Retrieval-Augmented Generation (RAG) pipeline.

\subsection{Data Preprocessing}

We process the raw data for each timestamp and based on the narrative context, we use regular expressions to extract the main textual content from the transcripts. We also retrieve essential metadata: \emph{language, source, duration, and collection information} for each transcript. This data is then structured to facilitate further analysis and model training. 

Preprocessed raw data is then structured into two key columns: "page content" and "metadata." This tabular format aligns with the document schema required by Langchain. The "page content" holds contextual data for fine-tuning and the RAG pipeline, while "metadata" ensures proper categorization by language and content. This cleaned data is then utilized for QA pair extraction and RAG-based evaluations.

\subsection{Question-Answer (QA) Pair Extraction}

We use Self-Instruct \cite{wang2023selfinstructaligninglanguagemodels} to create instruction-output pairs, i.e., question and answer pairs using a cloud LLM endpoint \footnote{https://console.groq.com/}. We then provide textual documents, corresponding to each transcript and a prompt to the LLM to generate QA pairs. Our prompt provides specific requirements to guide the model to generate detailed, conversational Alpaca-style Q\&A pairs \cite{taori2023alpaca}, ensuring comprehensive coverage of the transcript without repetition. Our dataset contains approximately 64,000 QA pairs in total. Table \ref{tab:qa-data-statistics} shows the total number of pairs generated for each language. This dataset is shared publically in two releases, one for English\footnote{https://huggingface.co/datasets/RedHenLabs/qa-news-2016} and another for other European languages\footnote{https://huggingface.co/datasets/RedHenLabs/news-euro-2016}. Column 2 shows the number of QA pairs used for fine-tuning, while the last column lists the samples in the evaluation set. Additionally we also share a vector database\footnote{https://github.com/WhissleAI/Multimodal-RAG/releases/tag/vectordb-2016-01} created using our one year of news transcripts

\begin{table}
\centering
\begin{tabular}{|l|l|l|}
\hline
Language & \#Fine-tune & \#Evaluation \\ \hline
English  &   43,676    &      399        \\ \hline
French   &    5209     &       60       \\ \hline
Spanish  &    5519     &       60       \\ \hline
German   &    4748     &         60     \\ \hline
Portuguese   &  4454       &       60       \\ \hline
\end{tabular}
\caption{Question-Answer pairs data statistics}
\label{tab:qa-data-statistics}
\end{table}


\section{Method}
We describe method we follow to fine-tune the pretrained LLM checkpoint using QA pairs. We then expand on template formation required for inferece. We conclude this section by describing RAG pipeline to give verifiable contextualized answers.

\subsection{Base Model}

We evaluate multiple off-the-shelf models, including Gemma-7B, Gemma-2B, Llama-3-8B, and Phi-3-mini-4K. Results in Table \ref{tab:model-comparison} indicate that the Phi-3-mini-4K model gives best results overall. We found that the Phi-3-mini-4K model \cite{abdin2024phi} is most suitable for multi-lingual query understanding and response generation. This Phi-3-mini-4K model has a 3.8 billion parameter model with a 4096 context window length, originally trained with over 3.3 trillion tokens. A standout feature of Phi-3-mini-4K is its ability to be quantized to 4 bits, reducing its memory footprint to approximately 1.8 GB. Notably, the model achieves a processing speed of 8 to 12 tokens per second using a single T4 GPU with 3-4 GB VRAM required for inference. 

\subsection{Fine-tuning and Inference}

For each QA pair, we use the question as an instruction with the system prompt \emph{You should act like a news reporter} and fine-tune the model to generate answers using cross-entropy loss. We employ QLoRA adapters for parameter-efficient fine-tuning, significantly reducing memory and computation needs through double quantization. This reduces trainable parameters by 99.52\%, bringing the model from 3.72 billion to 17.8 million parameters, enhancing scalability. To optimize training further, we use FlashAttention-2 with bfloat16 precision, accelerating memory-efficient attention and minimizing computational bottlenecks. bfloat16 balances memory efficiency with precision, allowing large models to be trained on consumer-grade hardware while maintaining high-quality inference.

Instruct-based fine-tuning also requires a structured prompt template with control tokens (e.g., user and assistant), ensuring the model interprets context and generates relevant responses. Table \ref{tab:prompt_template} shows an example of such a template.

\begin{table}
\centering
\begin{tabular}{|l|}
\hline
\texttt{<|user|>} \\
Act as a news reporter and answer the question: \\
Input: What is the most common side effect of taking Chantix? \\
\texttt{<|end|>} \\
\texttt{<|assistant|>} \\
\hline
\end{tabular}
\caption{Sample Prompt Template}
\label{tab:prompt_template}
\end{table}

\subsection{RAG: Constructing RAG with an LLM}
We develop a Retrieval-Augmented Generation (RAG) pipeline using LangChain\footnote{https://www.langchain.com/}, incorporating metadata for efficient source mapping. Transcripts are split into chunks, then vectorized using a pretrained sentence encoder, MPNET~\cite{song2020mpnet}, and stored with metadata in a \textit{Qdrant}\footnote{https://qdrant.tech/} database.

At inference, the query \( q \) is converted into a vector \( \mathbf{v}_q \) via MPNET. The transcript \( d \) is divided into chunks, each represented as a vector \( \mathbf{v}_{d_1}, \mathbf{v}_{d_2}, \dots, \mathbf{v}_{d_N} \), where \( N \) is the number of chunks. Cosine similarity between \( \mathbf{v}_q \) and each chunk embedding \( \mathbf{v}_{d_i} \) is calculated as:

\[
S_i = \frac{\mathbf{v}_q \cdot \mathbf{v}_{d_i}}{\|\mathbf{v}_q\| \|\mathbf{v}_{d_i}\|}, \quad i = 1, 2, \dots, N
\]

The top 4 chunks with the highest similarity are selected as additional context for the fine-tuned large language model (LLM). The LLM then generates the final answer \( A \) using the query \( q \) and the retrieved context.

\begin{table*}
\centering
\begin{tabular}{|l|c|c|c|c|c|c|}
\hline
(0 Shot) & News-reporter-3b & News-reporter-euro-3b & Phi-3-mini-4k & Gemma-7b-it & Llama-2-7B & Mistral-7B-Instruct-v0.2 \\
\hline
MMLU & 69.49 & 69.67 & \textbf{69.90} & 64.3 & 45.3 & 59.02 \\
\hline
ARC\_C & 56.40 & \textbf{57.51} & 56.14 & 53.2 & 45.9 & 55.89 \\
\hline
Winogrande & \textbf{74.19} & \textbf{74.19} & 73.24 & 68.03 & 69.5 & 73.72 \\
\hline
Truthfulqa & 50.43 & 43.45 & \textbf{66.46} & 44.18 & \textbf{57.4} & 53.00 \\
\hline
\end{tabular}
\caption{Comparison of language model performance across various benchmarks}
\label{tab:model-comparison}
\end{table*}
\section{Experiments \& Results}
We evaluate our fine-tuned model on standard LLM benchmarks and also on our evaluation set.

\subsection{Open LLM benchmark evaluation}

Table \ref{tab:model-comparison} shows results for our model on few standard open LLM benchmarks. \emph{News-reporter-3b} is the model fine-tuned only for English QA pairs while \emph{News-reporter-euro-3b} represents model fine-tuned for 5 languages, including English. Results show our models achieve competitive performance on the MMLU \cite{hendrycks2020measuring} benchmark, which tests a wide array of general knowledge topics. Fine-tuned models also outperforms other off-the-shelf models Gemma-7B and Llama-3-8B on MMLU benchmark. 

Our European model shows best results for \emph{ARC\_C} \cite{clark2018think} which measures reasoning ability of the model. It achieves best results for Winogrande \cite{sakaguchi2021winogrande} benchmark, which measures common-sense ability of the model. This highlights that news transcripts dataset is appropriate to improve reasoning and common-sense ability of LLM models. However, our results indicate a significant decline in performance for TruthfulQA\cite{lin2021truthfulqa} dataset. This indicates that model's performance degrades when it comes to factual questions about finance, health and law. 

\subsection{Retrieval and Generation Evaluation}

For RAG evaluation, we choose a set of metrics to assess both the retrieval and generation aspects of the pipeline. We use QA pairs from our evaluation set discussed in \ref{tab:qa-data-statistics}. We use RAGAS \cite{es2023ragas} to measure performance along four dimensions, namely:

\begin{itemize}
    \item Context Recall (CR): Extent of alignment between retrieved context and ground truth.
    \item Context Precision (CP): Presence and ranking of relevant ground-truth items in retrieved contexts.
    \item Answer Correctness (AC): Alignment of the generated answer with the ground truth.
    \item Answer Relevance (AR): Relevance and conciseness of the generated answer.
\end{itemize}

\begin{table}[htb]
\centering
\scalebox{0.8}{
\begin{tabular}{|l|l|c|c|c|c|}
\hline
\textbf{Model} & \textbf{Setting} & \textbf{CR} & \textbf{CP} & \textbf{AC} & \textbf{AR} \\ \hline
Phi-3-mini-4k & no rag & -      & -      & 0.3696  & 0.5431 \\ \hline
Phi-3-mini-4k & 1month rag & 0.5732 & 0.7287 & 0.4457  & 0.478 \\ \hline
News-reporter-3b      & no rag & -      & -      & 0.4867  & {\bf 0.828}  \\ \hline
News-reporter-3b      & 1month rag & {\bf 0.5764} & {\bf 0.7304} & {\bf 0.5111}  & 0.7519 \\ \hline
\end{tabular}
}
\caption{Performance Comparison of Different Models and Settings}
\label{tab:performance}
\end{table}

Results in Table \ref{tab:performance} show that our fine-tuned model shows significant improvement over off-the-shelf model. This highlights that our model is better at generating answers which are better contextualized, precise, correct and relevant for questions related to broadcast news.

We observe significant performance gains for both the off-the-shelf and fine-tuned models when using RAG with a vector database of broadcast news. This improvement was consistent, except in answer relevancy, where RAG showed a slight decrease compared to the fine-tuned model. We attribute this to the fine-tuned model's training on similar data, which already provided a substantial advantage over the off-the-shelf model. These findings are consistent for both English and other language evaluation datasets.

\section{Observation and Discussion}

\subsection{News related MMLU Topics}

The news-reporter-3b model generally performs better than the Phi-3-mini-4k in US foreign policy but scores slightly lower in public relations and miscellaneous topics. The lower scores in these areas may be due to the news-reporter-3b model's specialized focus on news content, which may not fully capture broader or diverse topic nuances.

\begin{table}[htb]
\centering
\scalebox{0.8}{
\begin{tabular}{|l|c|c|c|}
\hline
\textbf{MMLU Metric} & \textbf{Phi-3-mini-4k} & \textbf{news-reporter-3b} & \textbf{news-reporter-euro-3b} \\
\hline
Public relations & \textbf{71.82} & 70.91 & 68.18 \\
\hline
US\_foreign policy & 85.00 & \textbf{86.00} & \textbf{87.00} \\
\hline
Sociology & \textbf{86.07} & \textbf{86.07} & \textbf{86.07} \\
\hline
Miscellaneous & \textbf{83.78} & 82.50 & 82.12 \\
\hline
\end{tabular}
}
\caption{Comparison of General MMLU benchmark}
\end{table}

\subsection{Comparison of Models with RAG Enhancement}

\textbf{Question:} What is the current situation of Iran after the lifting of sanctions?

\begin{enumerate}
    \item \textbf{Off-the-Shelf Model:} \\
    Iran’s economy has improved, especially in the oil and gas sector, with increased exports and foreign investment. However, challenges like high unemployment and inflation persist. The government is working on reforms to address these issues.

    \item \textbf{Our Model with RAG:} \\
    Iran is becoming more prominent on the global stage after sanctions were lifted. A nuclear agreement has been reached, and talks are ongoing with international agencies to ensure compliance. Progress and stability in the region are key goals, but continuous efforts are required to meet conditions.

    \item \textbf{Ground Truth:} \\
    While nuclear-related sanctions have been lifted, others remain, such as those related to Iran’s support for terrorism. Economic issues, including high unemployment and a struggling oil industry, persist.

\end{enumerate}

Our fine-tuned RAG model provides a more detailed and nuanced answer, capturing recent developments better than the off-the-shelf model.

\section{Conclusions}

We create and share a high quality multilingual QA pair dataset gathered from multiple news channels. We provide an efficient method to fine-tune an existing model to understanding queries, contextualize via retrieving relevant news transcripts and  generate appropriate answers related to news. Our results show significant improvement in LLM's performance to answer questions related to broadcast news. We also show that broadcast news is suitable for LLMs to improve on few news relevant open LLM benchmarks.

\section{Future work}

This is a work in progress as we plan to extend our datasets to include a corpus of news transcripts covering a period longer than a year. This will help our \emph{News-reporter-3B} LLM improve further in answering queries related to broadcast news.

In an era when fake news spreads rapidly, we aim to build methods to apply guardrails that provide reliable and accurate answers based on verified TV sources. We also plan to incorporate information from audio and visual cues into our context dataset to enhance our ability to answer queries more effectively.

\bibliographystyle{IEEEbib}
\bibliography{refs}
\end{document}